\pdfoutput=1
\documentclass[11pt]{article}

\usepackage{acl}

\usepackage{times}
\usepackage{latexsym}
\usepackage{booktabs}
\usepackage{amsmath}

\usepackage[T1]{fontenc}
\usepackage[utf8]{inputenc}

\usepackage{microtype}

\usepackage{inconsolata}

\usepackage{graphicx}

\title{Language Barriers: Evaluating Cross-Lingual Performance of CNN and Transformer Architectures for Speech Quality Estimation}

\author{
 \textbf{Wafaa Wardah\textsuperscript{1,2}},
 \textbf{Tuğçe Melike Koçak Büyüktaş\textsuperscript{1}},
 \textbf{Kirill Shchegelskiy\textsuperscript{1}},\\
 \textbf{Sebastian Möller\textsuperscript{1,2}},
 \textbf{Robert P. Spang\textsuperscript{1}},
\\
\\
 \textsuperscript{1}Quality and Usability Lab, Technische Universität Berlin, Germany,\\
 \textsuperscript{2}Deutsches Forschungszentrum für Künstliche Intelligenz,\\Speech and Language Technologies, Berlin, Germany
\\
 \small{
   \textbf{Correspondence:} \href{mailto:wafaa.wardah@tu-berlin.de}{wafaa.wardah@tu-berlin.de}
 }
}

\begin{document}
\maketitle
\begin{abstract}
    Objective speech quality models aim to predict human-perceived speech quality using automated methods. However, cross-lingual generalization remains a major challenge, as Mean Opinion Scores (MOS) vary across languages due to linguistic, perceptual, and dataset-specific differences. A model trained primarily on English data may struggle to generalize to languages with different phonetic, tonal, and prosodic characteristics, leading to inconsistencies in objective assessments.
    
    This study investigates the cross-lingual performance of two speech quality models: NISQA, a CNN-based model, and a Transformer-based Audio Spectrogram Transformer (AST) model. Both models were trained exclusively on English datasets containing over 49,000 speech samples and subsequently evaluated on speech in German, French, Mandarin, Swedish, and Dutch. We analyze model performance using Pearson Correlation Coefficient (PCC) and Root Mean Square Error (RMSE) across five speech quality dimensions: coloration, discontinuity, loudness, noise, and MOS.
    
    Our findings show that while AST achieves a more stable cross-lingual performance, both models exhibit noticeable biases. Notably, Mandarin speech quality predictions correlate highly with human MOS scores, whereas Swedish and Dutch present greater prediction challenges. Discontinuities remain difficult to model across all languages. These results highlight the need for more balanced multilingual datasets and architecture-specific adaptations to improve cross-lingual generalization.
\end{abstract}

%%%%%%%%%%%%%%%%%%%%%%%%%%%%%%%%%%%%%%%%%%%%%%%%%%%%%%%%%%%%%%%%%
\section{Introduction}
%%%%%%%%%%%%%%%%%%%%%%%%%%%%%%%%%%%%%%%%%%%%%%%%%%%%%%%%%%%%%%%%%

    Objective speech quality models aim to predict perceptual speech quality across different languages. However, achieving cross-lingual generalization remains a complex challenge due to linguistic and perceptual variations, systematic biases in subjective ratings, and inconsistencies across datasets. Mean Opinion Scores (MOS), which serve as the ground truth for model training, can vary significantly by language, introducing systematic distortions that hinder generalization when models are primarily trained on monolingual datasets. For instance, tonal languages like Mandarin may exhibit greater cross-lingual stability due to their reliance on pitch contours rather than segmental phonemes~\cite{tang2020acoustic}. Additionally, Mandarin speech is perceived more robustly in noisy conditions, with neural evidence suggesting that tonal cue processing remains stable even under degraded audio~\citep{song2020increased}.
    
    Existing approaches have explored various architectures for speech quality prediction. Convolutional Neural Network (CNN)-based models, such as NISQA~\citep{mittag2021nisqa}, extract spectral features and have demonstrated strong correlations with MOS. However, they struggle to capture long-range dependencies in speech, which are essential for modeling complex degradations over time. In contrast, Transformer-based models, such as the Audio Spectrogram Transformer (AST)~\citep{gong2021ast}, leverage self-attention to model these dependencies more effectively. Despite their success in speech and audio processing tasks, their efficacy in cross-lingual speech quality assessment remains largely unexplored.  
    
    Addressing cross-lingual biases in objective speech quality models requires careful dataset design and evaluation methodologies. ITU-T Rec. P.1401~\citep{rec1401p} suggests mitigating discrepancies in MOS distributions through diverse multilingual datasets, scaling functions (e.g., polynomial mappings) to align subjective scores while preserving rank order, and ensuring that listening tests incorporate native speakers. However, achieving a fully controlled cross-lingual evaluation is challenging. Ideally, a comparative analysis would require datasets that differ only in language while remaining identical in all other aspects—such as recording quality, introduced degradations, and subjective rating conditions. This would involve speech samples recorded at uniform quality (e.g., 48 kHz), processed with identical degradation pipelines, and evaluated by native listeners under consistent rating protocols.  
    
    Unfortunately, such standardized datasets are currently unavailable. Instead, we utilize a diverse dataset compiled for the ITU-T P.SAMD initiative, which includes speech samples across multiple languages, raters, environments, and degradation types. While this dataset provides valuable insights into cross-lingual variability, it suffers from inconsistencies in degradation conditions, as data collection and processing were conducted independently across different laboratories. To ensure a meaningful comparison, we restrict our analysis to languages with at least 1,000 samples, as summarized in Table~\ref{tab:different_dbs}. Our final dataset comprises over 49,000 training samples, with Dutch being the smallest subset (1,035 samples). While this dataset is not strictly comparable across languages, we hypothesize that the extensive sample size per language offsets some of the inconsistencies inherent to the data.  
    
    This study systematically evaluates the factors influencing cross-lingual generalization in speech quality models. Specifically, we examine:
    \begin{itemize}
        \item \textbf{Source Material:} The impact of speaker characteristics, recording environments, and linguistic variations.
        \item \textbf{Degradations:} Differences in speech distortion types and their severity across datasets.
        \item \textbf{Model Architecture:} Comparing CNN-based (NISQA) and Transformer-based (AST) approaches.
        \item \textbf{Subjective Evaluations:} Variability in MOS ratings across linguistic groups.
        \item \textbf{Dataset Composition:} The influence of dataset structure, including single-speaker vs. conversational speech.
    \end{itemize}
    
    By analyzing these factors, we aim to provide deeper insights into how architectural differences between CNNs and Transformers influence speech quality modeling, the extent to which language-specific biases affect model predictions, and strategies to enhance the cross-lingual robustness of objective speech quality assessment models.

%%%%%%%%%%%%%%%%%%%%%%%%%%%%%%%%%%%%%%%%%%%%%%%%%%%%%%%%%%%%%%%%%
\section{Dataset}
%%%%%%%%%%%%%%%%%%%%%%%%%%%%%%%%%%%%%%%%%%%%%%%%%%%%%%%%%%%%%%%%%
    We used a diverse set of databases containing over 49,000 English speech samples for training, ensuring coverage of various distortions. For evaluation, we employed over 70,000 speech samples across six languages (German, French, Mandarin, Swedish, and Dutch) with subjective MOS ratings. To ensure robustness, we only included languages with at least 1,000 samples (Table~\ref{tab:different_dbs}). Since not all sub-dimensions were rated subjectively, objective scores from ITU-T P.863 / POLQA~\citep{rec8632} were used to supplement missing ratings.
    
    \begin{table}[h]
        \centering
        \begin{tabular}{lcccc}
            \toprule
            \textbf{Language} & \textbf{DBs} & \textbf{Conditions} & \textbf{Samples} \\
            \midrule
            \textbf{ENG$_{tr}$} & 34   & 10k & 49062 \\
            \midrule
            \textbf{ENG$_{te}$}  & 3    & 2295  & 28914 \\
            \midrule
            \textbf{DE}        & 26   & 125   & 27887 \\
            \textbf{SE}        & 7    & 234   & 3442  \\
            \textbf{FR}        & 10   & 56    & 3299  \\
            \textbf{MAN}       & 4    & 60    & 3072  \\
            \textbf{NL}        & 6    & 59    & 1035  \\
            \bottomrule
        \end{tabular}
        \caption{Details about the number of DBs, different conditions, total samples, and the ratio of samples per condition per language. Training (ENG$_{tr}$) and Test (ENG$_{te}$) sets used for the model creation are listed.}
        \label{tab:different_dbs}
    \end{table}

\section{Models}
    For this experiment, we trained two models on the dataset described above: the open-sourced NISQA model and a new proposed model based on the Audio Spectrogram Transformer (AST). 

    \subsection{NISQA}
\begin{figure}[h]
            \centering
            \includegraphics[width=1.0\linewidth]{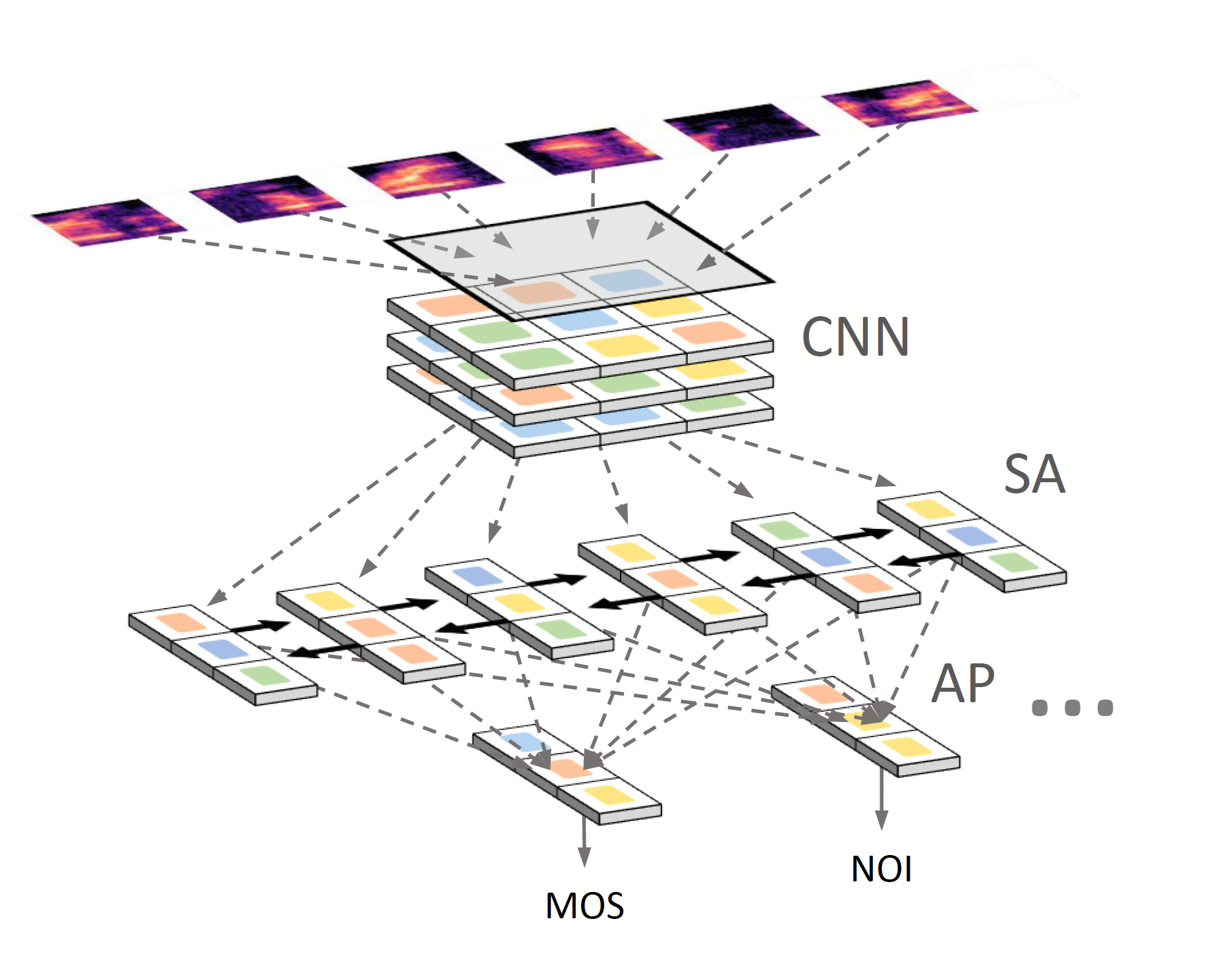}  % Adjust width as needed
            \caption{NISQA model architecture}
            \label{fig:model_nisqa}
        \end{figure}

        NISQA is a CNN/Mel-Spectrogram-based model that is openly available (see figure \ref{fig:model_nisqa}). The model was not pre-trained and was exclusively trained on the training dataset for this work. Training parameters included up to 500 epochs with an early stopping policy of 20 epochs, based on overall Pearson Correlation Coefficient (PCC) and Root Mean Square Error (RMSE) scores obtained on the test set after each epoch. Additional parameters were a block size of 100, ADAM optimizer with a learning rate of 0.001, and a learning rate patience of 15. 

    \subsection{AST-based model}
\begin{figure}[h]
            \centering
            \includegraphics[width=1.0\linewidth]{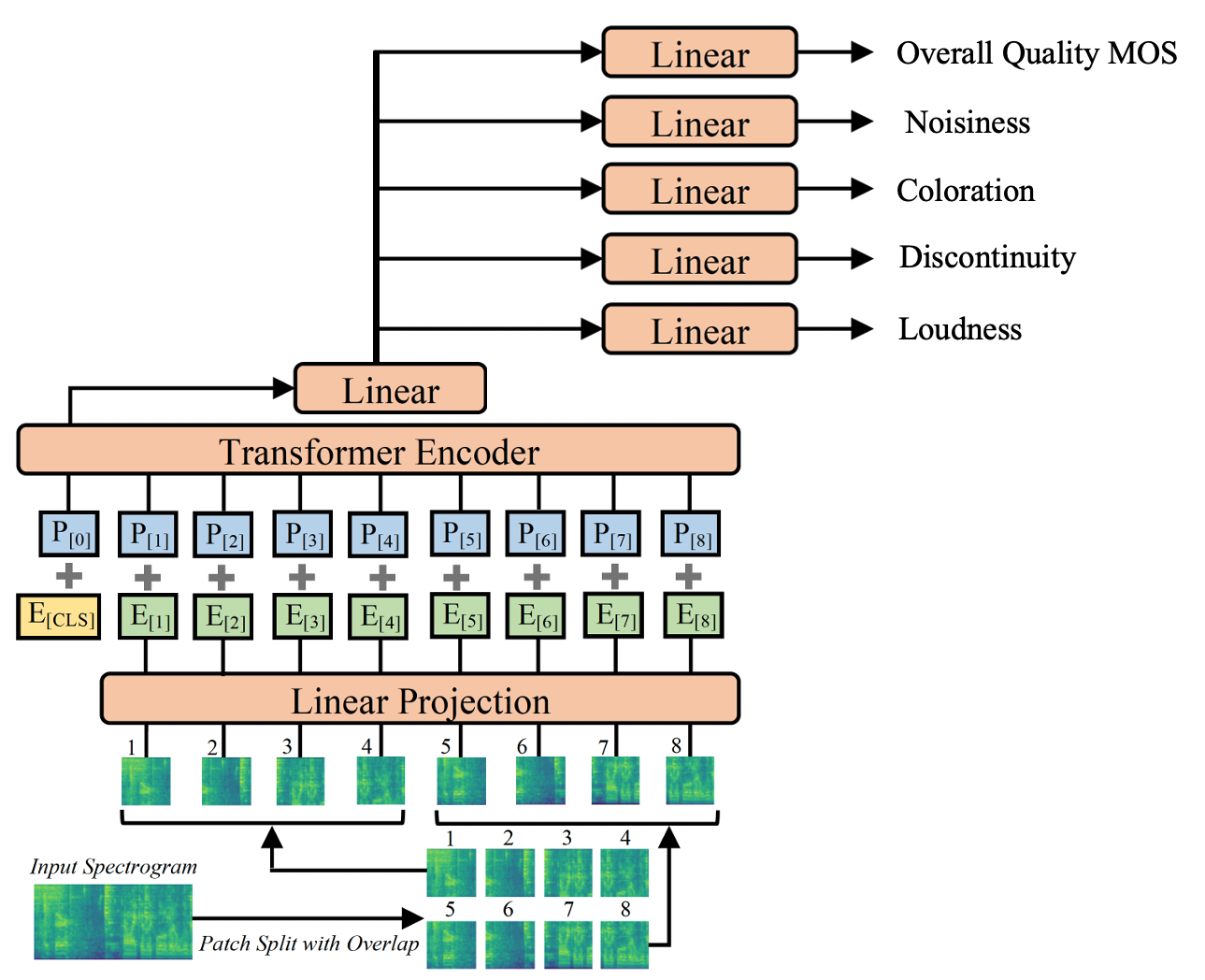}  % Adjust width as needed
            \caption{AST-based model architecture}
            \label{fig:model_ast}
        \end{figure}

        We propose a transformer-based architecture that leverages the open-sourced AST model. The feature extraction pipeline follows the original design proposed by the AST authors, with some modifications. While the original implementation used waveforms sampled at 16 kHz, our approach processes audio sampled at 48 kHz for higher fidelity.nLog-Mel spectrograms are computed from the input waveform using a 25 ms window size, a 10 ms hop size, and 128 Mel frequency bins. For audio clips shorter than 12 seconds, an attention mask is applied to handle variable-length sequences. The resulting spectrogram is divided into overlapping patches of size 16 × 16, with overlaps applied in both the time and frequency dimensions. Each patch is flattened into a 1D vector of size 768 through a linear projection layer, referred to as the patch embedding layer. A classification (CLS) token is prepended to the sequence to facilitate downstream tasks. To preserve the 2D spatial structure of the spectrogram, a learnable positional embedding of size 768 is added to the patch embeddings, enabling the model to effectively capture temporal and spectral relationships within the input features. We implemented downstream multitasking by feeding the output of the AST model into five separate linear layers, each dedicated to predicting a specific aspect of speech quality: the overall MOS score and the four perceptual dimensions (see Figure \ref{fig:model_ast}). The loss for each task is computed separately and backpropagated, ensuring that each linear layer specializes in its respective dimension. Meanwhile, the AST backbone remains fully shared and is updated across all tasks, allowing it to learn a more generalized audio representation while preserving task-specific specialization. The model was trained using the Mean Squared Error (MSE) loss function, optimized with the ADAM optimizer and a learning rate of 1e-6.

        \noindent\textbf{Pre-trained version} \\ 
        Since transformer models start seeing the benefits of the architecture only after a very large quantity of training data, we explored the pre-trained AST weights and along with a version that was not pre-trained. The pre-trained version of the AST model leverages cross-modality transfer learning by initializing its weights using pre-trained weights from a Vision Transformer (ViT) model originally designed for image-based tasks. The AST was further fine-tuned for audio classification on the AudioSet dataset, enabling it to learn task-specific features for audio signal processing. 
        
        AudioSet is a large-scale dataset containing over 2 million 10-second audio clips sourced from YouTube, covering 632 sound event categories. While AudioSet includes a diverse range of spoken languages, it is not specifically designed for linguistic analysis, as its primary focus is on general audio event detection. The dataset provides English-language labels for sound events (e.g., "speech," "car engine," "music") but does not annotate the specific language spoken in the clips. Consequently, the pre-training of AST on AudioSet allows it to learn broad audio features, but it does not inherently incorporate linguistic biases or language-specific speech processing. A detailed description of this pre-training can be obtained from the original paper [1].

    \subsection{Score Calibration}
        To align predicted MOS scores with human subjective ratings across datasets, we applied a third-order polynomial mapping, following ITU-T P.1401 recommendations~\citep{rec1401p}. This transformation reduces systematic prediction errors and harmonizes scores across datasets with different rating distributions.

%%%%%%%%%%%%%%%%%%%%%%%%%%%%%%%%%%%%%%%%%%%%%%%%%%%%%%%%%%%%%%%%%
\section{Results}
%%%%%%%%%%%%%%%%%%%%%%%%%%%%%%%%%%%%%%%%%%%%%%%%%%%%%%%%%%%%%%%%%

    In this section, we present the evaluation results of the cross-lingual performance of the NISQA (CNN-based) and AST (Transformer-based) speech quality assessment models. We report Pearson Correlation Coefficient (PCC) and Root Mean Square Error (RMSE) values for five quality dimensions: coloration (Col), discontinuity (Dis), loudness (Loud), noise (Noi), and mean opinion score (MOS). 

    \subsection{Cross-Lingual Performance of NISQA}
    
        The PCC results for the NISQA model (Table~\ref{tab:pcc_values_language_comparison_eng_nisqa}) indicate that performance varies across languages, with the highest correlation observed for noise (0.927 for Mandarin) and the lowest for loudness (0.511 for French). The overall PCC range across languages is largest for loudness (0.12) and noise (0.29), suggesting that these dimensions are more susceptible to language-based variability. The PCC values for MOS are highest for Mandarin (0.835) and French (0.826), while notably lower for Swedish (0.673), indicating potential inconsistencies in subjective ratings across languages.

        \begin{table}[h]
            \centering
            \begin{tabular}{lccccc}
                \toprule
                \textbf{PCC} & \textbf{Col} & \textbf{Dis} & \textbf{Loud} & \textbf{MOS} & \textbf{Noi} \\
                \midrule
                \textbf{ENG} & 0.759 & 0.363 & 0.586 & 0.718 & 0.879 \\
                \midrule
                \textbf{MAN} & 0.799 & 0.815 & 0.769 & 0.835 & 0.927 \\
                \textbf{FR}  & 0.764 & 0.794 & 0.511 & 0.826 & 0.856 \\
                \textbf{DE}  & 0.665 & 0.728 & 0.574 & 0.800 & 0.808 \\
                \textbf{SE}  & 0.641 & 0.698 & 0.544 & 0.673 & 0.798 \\
                \textbf{NL}  & 0.622 & 0.726 & 0.622 & 0.798 & 0.641 \\
                \midrule
                \textbf{Range} & 0.18 & 0.12 & 0.26 & 0.16 & 0.29 \\
                \bottomrule
            \end{tabular}
            \caption{PCC results of the language comparison of the English trained NISQA model.}
            \label{tab:pcc_values_language_comparison_eng_nisqa}
        \end{table}
        
        Similarly, the RMSE results (Table~\ref{tab:rmse_values_language_comparison_eng_nisqa}) reveal that errors in prediction vary across languages. The largest RMSE range is observed for noise (0.19), followed by discontinuity (0.15), indicating that the model struggles most with accurately predicting these dimensions in different languages. The MOS RMSE results highlight increased prediction errors for Mandarin (0.568) and Swedish (0.627), suggesting that the model exhibits higher deviations from ground-truth MOS in these languages compared to English (0.280).  

        \begin{table}[h]
            \centering
            \begin{tabular}{lccccc}
                \toprule
                \textbf{RMSE} & \textbf{Col} & \textbf{Dis} & \textbf{Loud} & \textbf{MOS} & \textbf{Noi} \\
                \midrule
                \textbf{ENG} & 0.302 & 0.357 & 0.326 & 0.280 & 0.480 \\
                \midrule
                \textbf{MAN} & 0.386 & 0.439 & 0.282 & 0.568 & 0.289 \\
                \textbf{FR}  & 0.361 & 0.429 & 0.295 & 0.518 & 0.376 \\
                \textbf{DE}  & 0.406 & 0.460 & 0.251 & 0.499 & 0.385 \\
                \textbf{SE}  & 0.333 & 0.411 & 0.297 & 0.627 & 0.319 \\
                \textbf{NL}  & 0.408 & 0.564 & 0.364 & 0.539 & 0.477 \\
                \midrule
                \textbf{Range} & 0.07  & 0.15  & 0.11  & 0.13  & 0.19  \\
                \bottomrule
            \end{tabular}
            \caption{RMSE results of the language comparison of the English trained NISQA model.}
            \label{tab:rmse_values_language_comparison_eng_nisqa}
        \end{table}

    \subsection{Cross-Lingual Performance of AST}
    
        The highest PCC for MOS is obtained for Mandarin (0.792) and French and Dutch (both 0.79), while the lowest is for Swedish (0.649). The model exhibits strong correlations for noise across most languages, with the highest observed for Mandarin (0.899). However, discontinuity yield significantly lower PCC values, similar to NISQA, suggesting that both models struggle in predicting subjective scores for this particular quality dimension. The PCC ranges in AST are highest for loudness (0.22), indicating variability in model robustness across different languages.  

        \begin{table}[h]
            \centering
            \begin{tabular}{lccccc}
                \toprule
                \textbf{PCC} & \textbf{Col} & \textbf{Dis} & \textbf{Loud} & \textbf{MOS} & \textbf{Noi} \\
                \midrule
                \textbf{EN}  & 0.814 & 0.359 & 0.605 & 0.701 & 0.875 \\
                \midrule
                \textbf{MAN} & 0.768 & 0.550 & 0.618 & 0.792 & 0.899 \\
                \textbf{SE}  & 0.631 & 0.651 & 0.601 & 0.649 & 0.768 \\
                \textbf{DE}  & 0.662 & 0.658 & 0.571 & 0.768 & 0.808 \\
                \textbf{FR}  & 0.732 & 0.728 & 0.496 & 0.790 & 0.827 \\
                \textbf{NL}  & 0.583 & 0.737 & 0.715 & 0.790 & 0.706 \\
                \midrule
                \textbf{Range} & 0.19  & 0.19  & 0.22  & 0.14  & 0.19  \\
                \bottomrule
            \end{tabular}
            \caption{PCC results of the language comparison of the English trained AST model.}
            \label{tab:pcc_dimension}
        \end{table}
    
        The RMSE results for AST (Table~\ref{tab:rmse_dimension}) follow a similar trend to those of NISQA, with the highest errors observed for MOS in Swedish (0.641) and Dutch (0.553). The variability in RMSE across languages remains notable, particularly for MOS (range: 0.19).

        \begin{table}[h]
            \centering
            \begin{tabular}{lccccc}
                \toprule
                \textbf{RMSE} & \textbf{Col} & \textbf{Dis} & \textbf{Loud} & \textbf{MOS} & \textbf{Noi} \\
                \midrule
                \textbf{EN}  & 0.263 & 0.363 & 0.308 & 0.290 & 0.489 \\
                \midrule
                \textbf{MAN} & 0.375 & 0.463 & 0.350 & 0.454 & 0.431 \\
                \textbf{SE}  & 0.323 & 0.418 & 0.252 & 0.641 & 0.329 \\
                \textbf{DE}  & 0.405 & 0.498 & 0.264 & 0.505 & 0.413 \\
                \textbf{FR}  & 0.331 & 0.442 & 0.260 & 0.525 & 0.333 \\
                \textbf{NL}  & 0.414 & 0.557 & 0.298 & 0.553 & 0.473 \\
                \midrule
                \textbf{Range} & 0.09  & 0.14  & 0.10  & 0.19  & 0.14  \\
                \bottomrule
            \end{tabular}
            \caption{RMSE results of the language comparison of the English trained AST model.}
            \label{tab:rmse_dimension}
        \end{table}

%%%%%%%%%%%%%%%%%%%%%%%%%%%%%%%%%%%%%%%%%%%%%%%%%%%%%%%%%%%%%%%%%
\section{Discussion}
%%%%%%%%%%%%%%%%%%%%%%%%%%%%%%%%%%%%%%%%%%%%%%%%%%%%%%%%%%%%%%%%%

    The results of this study highlight key challenges and insights regarding the cross-lingual generalization of objective speech quality assessment models. Specifically, both the CNN-based NISQA model and the Transformer-based AST model exhibit varying degrees of language dependency, with notable discrepancies across different quality dimensions. While the AST model generally achieves a more consistent performance across languages, certain language-dimension pairs continue to show substantial variability, raising important questions about dataset composition, model architecture, and the role of linguistic characteristics in speech quality prediction.
    
    \subsection{Cross-Lingual Differences and Dataset Challenges}
    
        Despite being trained solely on English data, both models exhibit strong MOS correlations across multiple languages. However, individual perceptual dimensions show notable variations, underscoring the complexity of speech quality perception. Prior studies have shown that overall speech quality is not solely determined by MOS but rather by an interplay of perceptual dimensions such as noisiness, coloration, and discontinuities \cite{wardah2024speechquality}. Our results reinforce this, as different languages exhibit distinct correlation patterns across these dimensions.  

        Mandarin, in particular, consistently stands out, achieving high PCC scores across most dimensions with both models. This trend aligns with prior research indicating that Mandarin's tonal characteristics and relatively stable spectral structure contribute to its robust performance in speech modeling~\citep{tang2020acoustic}. The spectral and temporal features governing Mandarin's speech quality degradations appear to be well captured, even by models trained exclusively on English data.

        \begin{figure}[h] % h: Place here, t: top, b: bottom, p: page
            \centering
            \includegraphics[width=1.0\linewidth]{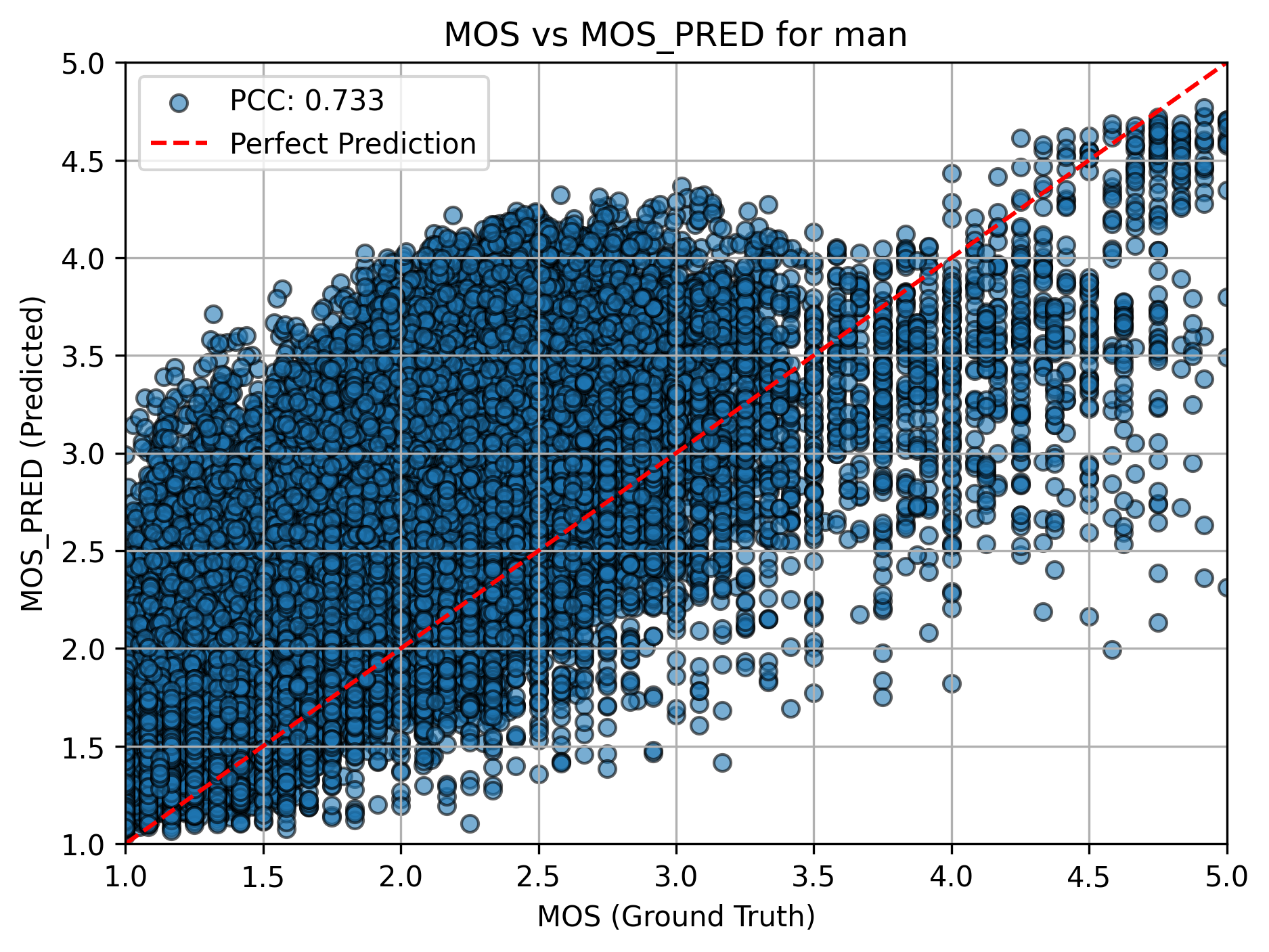} % Adjust width as needed
            \caption{MOS Prediction Scatterplot for Mandarin.} 
            \label{fig:example_image} % Refer to this figure using \ref{fig:example_image}
        \end{figure}
    
        Conversely, Swedish and Dutch represent significant challenges, particularly for MOS predictions. The AST model still ranks Swedish as the lowest-performing language for MOS, while Dutch continues to experience the worst performance in coloration and noisiness in both models. Strikingly, the discontinuity dimension demonstrates opposite trends for Mandarin, which yielded the highest PCC with NISQA but the lowest with AST. This observation suggests that different model architectures process linguistic features and degradations in distinct ways, which should be more closely examined in future studies.

        \begin{figure}[h] % h: Place here, t: top, b: bottom, p: page
            \centering
            \includegraphics[width=1.0\linewidth]{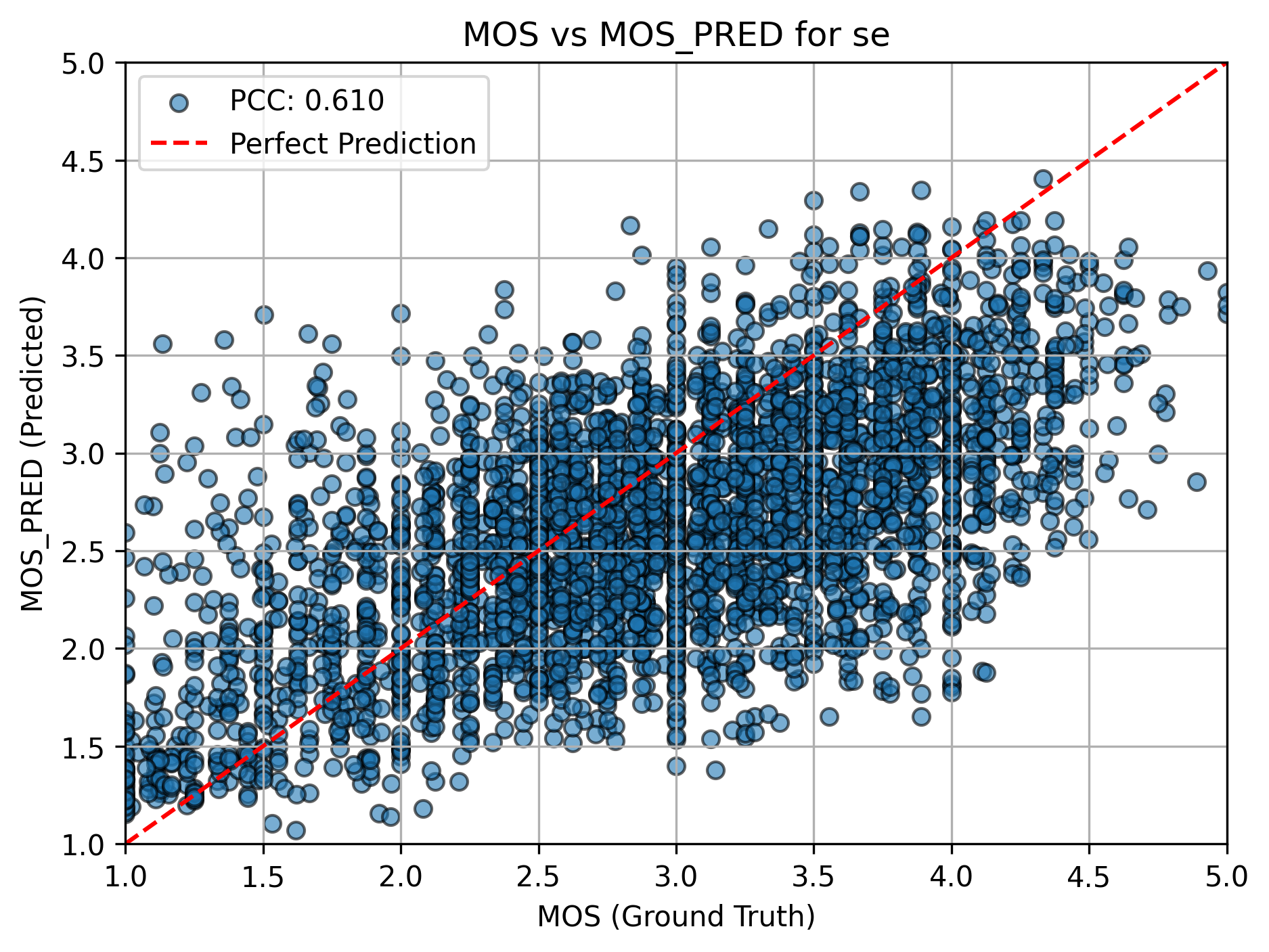} % Adjust width as needed
            \caption{MOS Prediction Scatterplot for Swedish.} 
            \label{fig:example_image} % Refer to this figure using \ref{fig:example_image}
        \end{figure}
    
        French exhibits the poorest PCC results for loudness in both models. Since this trend persists regardless of architecture, it is likely attributable to dataset-specific issues rather than an intrinsic linguistic effect. In general, our findings indicate that some language-dimension pairs produce consistent results across models, suggesting that dataset structures and sample balance may heavily influence performance. In contrast, cases where a language-dimension pair yields the best results in one model but the worst in another imply that model architectures differentially handle language-specific distortions.
    
    \subsection{Comparing Model Architectures}
    
        The AST model demonstrates a more stable performance across languages, with reduced PCC variability in most dimensions compared to NISQA. This aligns with expectations, given that Transformer-based architectures leverage self-attention mechanisms that capture long-range dependencies more effectively. However, the observed trade-offs in language-specific generalization indicate that the shift from CNN-based feature extraction to Transformer-based modeling does not uniformly mitigate all cross-lingual biases.
    
        The fact that Mandarin achieved the strongest results in coloration, loudness, and noisiness with NISQA, whereas discontinuity performed best in Dutch with AST, highlights fundamental differences in how these architectures perceive degradation artifacts. The AST model seems to handle Dutch speech distortions better in discontinuity and loudness but remains highly sensitive to noise and coloration variations in this language. These findings emphasize that architecture choices significantly impact speech quality estimation performance across languages.
    
        Interestingly, an artifact observed in the English validation set reveals a pronounced skew in the discontinuity dimension. Since validation in this study relied on objectively scored datasets (as all subjectively scored datasets were incorporated into training), this highlights a potential issue in how discontinuity are defined and assessed in automatic scoring algorithms. The lower discontinuity correlation in English suggests inconsistencies in objective degradation modeling, which merits further investigation.
    
    \subsection{Future Directions}
    
        These findings emphasize the need for future research aimed at refining model architectures and dataset structures to enhance cross-lingual robustness. Possible directions include:
        \textbf{Balanced Multilingual Datasets}: The discrepancies observed across languages suggest that dataset imbalances in degradation conditions and subjective ratings contribute to inconsistent performance. A concerted effort to develop high-quality, linguistically diverse datasets with standardized degradation pipelines is crucial.
        \textbf{Calibration Mechanisms}: Scaling functions and fine-tuning strategies tailored to specific linguistic traits may help mitigate biases introduced by dataset variability. Particularly, improving discontinuity detection across languages appears critical given its low PCC results across both models.
        \textbf{Reliance on Subjective Labels}: The observed skew in the English discontinuity dimension suggests that objective scoring methods may introduce unintended biases, leading to inconsistencies across languages. To ensure more reliable model evaluation, future work should prioritize subjective MOS labels over objective scores, thereby reducing the risk of error propagation from automated scoring models.

%%%%%%%%%%%%%%%%%%%%%%%%%%%%%%%%%%%%%%%%%%%%%%%%%%%%%%%%%%%%%%%%%

\bibliography{main}

% \appendix

% \section{Example Appendix}
% \label{sec:appendix}

% This is an appendix.

\end{document}